\documentclass[a4paper,twoside]{article}

\usepackage{epsfig}
\usepackage{subcaption}
\usepackage{calc}
\usepackage{amssymb}
\usepackage{amstext}
\usepackage{amsmath}
\usepackage{amsthm}
\usepackage{multicol}
\usepackage{pslatex}
\usepackage{apalike}
\usepackage{algorithm2e}
\usepackage{multirow}
\usepackage{adjustbox}
\usepackage[bottom]{footmisc}
\usepackage{SCITEPRESS}     

\begin{document}

\title{Abnormal Event Detection In Videos Using Deep Embedding}

\author{\authorname{Darshan Venkatrayappa}
\email{darsh.venkat@gmail.com}
}


\abstract{Abnormal event detection or anomaly detection in surveillance videos is currently a challenge because of the diversity of possible events. Due to the lack of anomalous events at training time, anomaly detection requires the design of learning methods without supervision. In this work we propose an unsupervised approach for video anomaly detection with the aim to jointly optimize the objectives of the deep neural network and the anomaly detection task using a hybrid architecture. Initially, a convolutional autoencoder is pre-trained in an unsupervised manner with a fusion of depth, motion and appearance features. In the second step, we utilize the encoder part of the pre-trained autoencoder and extract the embeddings of the fused input. Now, we jointly train/ fine tune the encoder to map the embeddings to a hypercenter. Thus, embeddings of normal data fall near the hypercenter, whereas embeddings of anomalous data fall far away from the hypercenter.}

\onecolumn \maketitle \normalsize \setcounter{footnote}{0} \vfill

\section{\uppercase{Introduction}}
\label{sec:introduction}
To ensure the safety and security of public spaces, there is a need for swift and precise detection of abnormal events, including incidents such as altercations or urgent situations like fires. Achieving this goal involves strategically placing surveillance cameras in various locations such as airports, malls, and public streets, resulting in a significant increase in the volume of video data. However, manually detecting these events, or anomalies, is an extremely meticulous task that often demands more manpower than is readily available. This challenge is exacerbated by the low probability of these abnormal events occurring, making manual detection a laborious and time-consuming endeavor. Consequently, there is an urgent requirement for automated systems capable of identifying rare or unusual incidents and activities within surveillance videos.

Defining anomalies can pose a significant challenge, primarily because their determination relies heavily on the specific context in which they occur. For example, a person crossing the road might be considered anomalous if it happens outside of a designated crosswalk. Furthermore, the precise definition of what constitutes an anomaly can often be vague and subject to interpretation. Different individuals may have varying opinions on whether activities like walking around on a subway platform should be categorized as normal or flagged as anomalous, potentially due to suspicions or different perspectives \cite{paper1}. These complexities pose significant obstacles when attempting to construct a supervised learning model for distinguishing anomalies from regular occurrences, mainly because abnormal events represent only a small fraction of the total dataset.

To address this challenge, a part of the computer vision community approaches the anomaly detection problem as an outlier detection task. They construct a normality model based on training data representing normal events and label deviations from this model as anomalous. In our work, we adopt a similar approach by mapping the embeddings of our hybrid architecture to a hypercenter. Consequently, embeddings of normal data cluster closely around the hypercenter, while those of anomalous data are positioned far away from it. This enables us to effectively identify anomalies within the data.

\section{\uppercase{RELATED WORK}}


In the realm of computer vision, various approaches have been proposed to tackle the issue of abnormal event detection. The choice of a specific approach often hinges on several factors, including the nature of the input data (whether it's sequential or non-sequential) \cite{Chalapathy}, the type of labels available (supervised, unsupervised, or semi-supervised), and the desired output format (binary values or normality scores). Broadly speaking, we can categorize these approaches into two primary groups: those designed for sequential data, such as videos, and those tailored for non-sequential data, like images. When dealing with sequential data, such as videos, the focus typically shifts towards techniques rooted in CNNs (Convolutional Neural Networks), RNNs (Recurrent Neural Networks), or LSTM networks (Long Short-Term Memory Networks). On the other hand, approaches geared toward non-sequential data, like images, tend to favor the utilization of CNNs or AEs (Auto-encoders).


Anomaly detection can be broadly categorized along two axes: supervised approaches and unsupervised or semi-supervised approaches . In supervised methods, a dataset is meticulously labeled to distinguish normal from abnormal instances, effectively transforming the problem into a conventional classification task. While supervised approaches \cite{GoernitzKRB13} are generally more effective than unsupervised ones, they demand a substantial number of annotations, which, in our specific context, are seldom obtainable for abnormal events. Conversely, unsupervised approaches do not depend on pre-existing labels and instead rely on the inherent characteristics of the dataset to pinpoint anomalous instances. These techniques typically operate on the premise that abnormal data points are sparsely distributed within the dataset, prompting the search for examples that exhibit substantial deviations from the overall data distribution. Unsupervised methods frequently employ dimensionality reduction techniques like PCA, auto-encoders \cite{Zhou}, or generative models to achieve this goal. The output of the anomaly detector can be either a binary value or a "normality" score. This can be done either globally on the signal studied, or locally to indicate the position of the anomaly. Generally binary approaches are based on thresholding a score and we consider here only approaches returning a score. Most methods calculate distances between the data to be tested and a central point of a "normality" sphere \cite{Ruff}; any point sufficiently far from this center is considered abnormal.

According to \cite{Kiran} abnormal event detection in videos can be categorized into three main approaches. Firstly, there are reconstruction-based methods that focus on reducing data dimensionality, often through techniques like PCA \cite{KudoMMT13,Wang2019} or auto-encoders \cite{Chalapathy,Sabokrou_2016,Hasan_2016,Akhriev}. These methods assume that anomalies are in-compressible and thus cannot be effectively reconstructed from low-dimensional projections. These methods demonstrate promising results when the anomaly ratio is fairly low. Although the reconstruction of anomalous samples, based on a reconstruction scheme optimized for normal data, tends to generate a higher error, a significant amount of anomalous samples could mislead the autoencoders to learn the correlations in the anomalous data instead \cite{Tangqing}.

 Secondly, prediction-based methods take a different approach by employing auto-regressive models or generative models to predict successive video frames based on prior frames. Anomalies are identified when these predictions deviate significantly from the actual frames. \cite{Zhao2017} thus uses a 3D autoencoder for anomaly detection. Its decoder is composed of two parts, one allowing the reconstruction of the input sequence and the other predicting the following sequence. Liu et al. \cite{Liu2018} train a frame prediction network by incorporating different techniques including gradient loss, optical flow, and adversarial training. Authors of \cite{Hu2016} uses Slow Feature Analysis (SFA) to detect anomalies. However, it's noteworthy that many prediction-based techniques may not fully exploit the temporal context and high-level semantic information of video anomalies \cite{Zhang2023}. Furthermore, these sequential predictions can be computationally intensive, and the learned representations may not be optimized for anomaly detection, as their primary objective revolves around sequential prediction rather than anomaly identification. 
 
 Finally, generative-based methods utilize models like VAEs and GANs to understand the distribution of "normal" examples, aiding in anomaly detection. Authors of \cite{FAN2020} have used VAEs for video anomaly detection. Although these variational approaches are able to generate various plausible outcomes, the predictions are blurrier and of lower quality compared to state-of-theart GAN-based models. 
Authors of \cite{Ravanbakhsh_2017} proposed to learn the generator as a reconstructor of normal events, and
hence if it cannot properly reconstruct a chunk of the input frames, that chunk is considered an anomaly. However, adversarial training is unstable. Without an explicit latent variable interpretation, GANs are prone to mode collapse as generator fails to cover the space of possible predictions by getting stuck into a single mode.

The drawback of these above methods is that they do not detect anomalies directly. They instead leverage proxy tasks for anomaly detection, e.g., reconstructing input frames or predicting future frames, to extract general feature representations rather than normal patterns. To address these issues, we exploit the one-class classification objective to map normal data to the center of hypersphere. Specifically, we minimize the distance to the center of the hypersphere such that normal samples are mapped closely to the center of the sphere.  We achieve this by training a hybrid architecture with a fusion of motion, depth and appearance features belonging to the normal data. The embeddings from the encoder of the hybrid architecture are mapped to the center of the hypersphere. During the test time the normal data are mapped near the hypercenter, where as the abnormal data are mapped away from the hypercenter. 

\section{METHODOLOGY}
\begin{figure*}
	\centering
	\includegraphics[width=17cm,height=6cm]{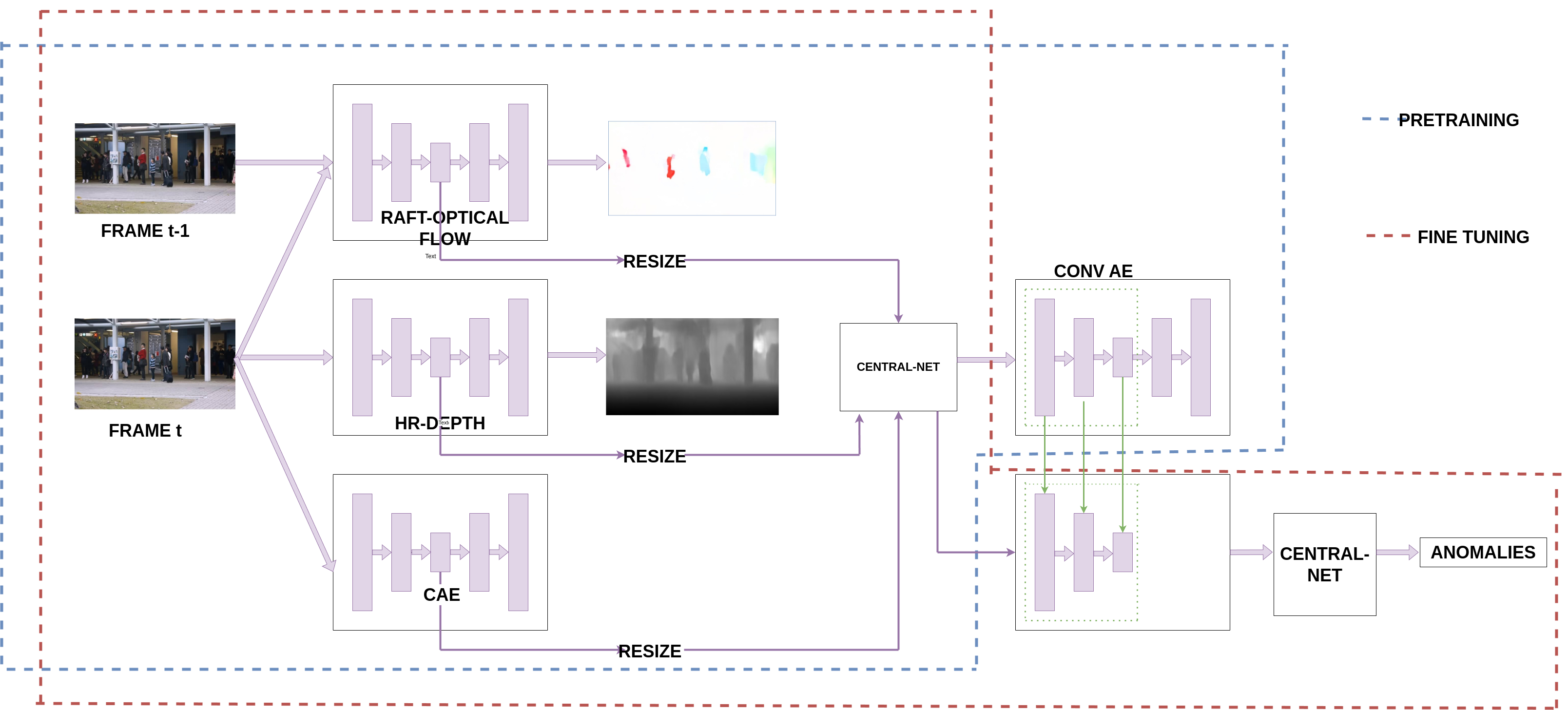}
	\caption{Overview of the method}
 \label{fig:Overview}
\end{figure*} 
Overview of our method is shown in Fig. \ref{fig:Overview}.  The proposed hybrid architecture is split in to 3 parts. 
\begin{itemize}
    \item Latent Feature Extraction.
    \item Feature Fusion.
    \item One class classification
\end{itemize}

\subsection{Latent Feature Extraction}
In our work we make use of 3 different input modalities : 
\begin{itemize}
    \item Depth maps : HR-Depth \cite{HR_DEPTH}.
    \item Optical flow : RAFT \cite{RAFT}.
    \item Appearance features : CAE(Convolution Auto encoder).
\end{itemize}
The architectures of these modalities are based on the encoder/decoder principle. We will use the outputs of the encoders to have a latent representation of each modality which will then be fused to detect anomalies. We use pre-trained models on modality-specific bases. More information about these modalities can be found in \cite{HR_DEPTH,RAFT}. Instead of Appearance features from the CAE, Latent features of Semantic maps from Mask-RCNN can also be used. In order to ensure that the modality extractor networks continue to fulfil this role, we decided to freeze their weights for learning. As the anomaly bases do not have any labelling on the modalities used, updating the modalities in an end-to-end network is made more difficult. A future extension of the work would be to use unsupervised cost functions to perform this task.

\subsection{Feature Fusion}
\begin{figure*}
	\centering
	\includegraphics[width=13cm,height=4cm]{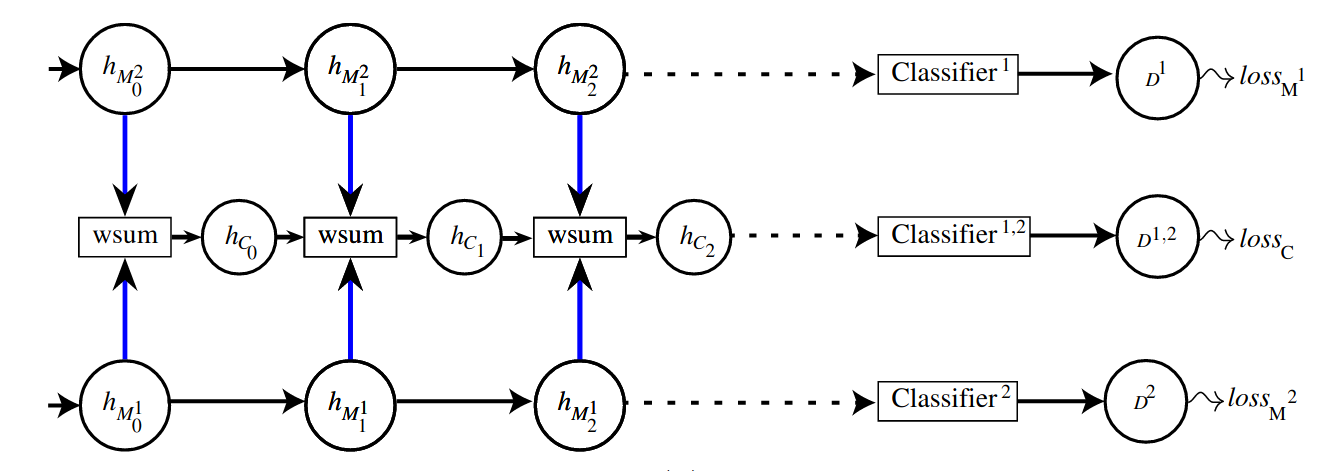}
	\caption{Architecture of fusion block}
 \label{fig:centralnet}
\end{figure*}

The optimal fusion of different modalities is a critical consideration, with the literature on multi-modal fusion offering numerous approaches but no consensus on the best fusion level. Instead, the ideal fusion position seems to be task-dependent, making it a parameter to be optimized like any other. Following the methodology proposed by the authors of \cite{CENTRAL-NET}, we implement feature fusion. As illustrated in Figure \ref{fig:centralnet}, the fusion block integrates features from various modalities through multiple branches. Each modality includes a standard neural network handling the modality. In addition, there is a central network that merges the modalities. At each layer of this central branch we combine via a weighted sum the values of the previous layer and the values of the layers of the same depth of the networks handling the modality. These fusion weights  are learned like the other parameters of the architecture. Thus the input to the fusion layer corresponds to the following equation:
$$h_{C_{i+1}} = \alpha_{C_i} h_{C_i} + \sum_{k=1}^m \alpha_{M_i^k} h_{M_i^k}.$$ 
Where $m$ is the number of modalities, $\alpha$ a learnable scalar, $h_{M_i^k}$ the hidden representation of each modality of depth $i$ and $h_{C_i}$ the central hidden representation. For each branch of the fusion block, we propose to use the  convolutional neural network. 

The encoder outputs of each modality are of different sizes so in order to process them in fusion block we need to align their dimensions. To do this, we use interpolation to obtain maps of the same size.

\subsection{One Class Classification}
The final stage of our approach is to use a one-class learning algorithm to detect anomalies. The goal of one-class learning is to build a classifier that identifies objects of the same nature as those presented during training. For this type of approach, the training set is only composed of the objects of interest/ normal data. Unlike traditional classification approaches, there is no attempt to distinguish between two or more classes with objects in each class. The focus here is on the class of interest and the aim is to delimit its boundaries. Historically, this type of classification has been studied on classifiers such as SVMs by seeking to identify the smallest hypersphere encompassing the training data. The authors of \cite{XuRYSS15} use this approach for anomaly detection by using auto-encoder. More recently, one-class learning has been extended to deep neural network \cite{Ruff}. In this article, we propose to use the \textit{One-Class Deep SVDD} approach of \cite{Ruff}.

The final stage is split in to two parts 1) Pre-training and 2) Fine tuning. In the pretraining stage we train a convolution autoencoder with the fused feature maps as the input. In the finetuning stage we use the encoder part of the pretrained CAE to extract the features and map the features to the center of the hypersphere such that the distance between the center of the hypersphere and the features are minimized. We use the following regularised loss function: 

$$loss_m(W,c) = \dfrac{1}{n} \sum_{i=1}^n \|\phi(x_i,W)-c\|^2 + \dfrac{\lambda}{2}\sum_{\ell=1}^L \|W^\ell\|^2_F,$$

where $W$ are the weights of the finetuned architecture, $x_i$ the $n$ examples of the mini-batch. $\phi$ is the inference of the network on the example $x_i$, for the weights $W$. $\lambda>0$ is the weight of the regularisation and $c$ the hypercentre of the reference distribution. We thus seek the same hypercentre $c$ surrounding the reference data set. In the case of learning the parameters of the fusion block, we use a cost function per network branch. Consequently, the learning process consists in solving the following problem:

$$\arg\min_{W,c_f,c_1,\cdots,c_m} loss_f(W,c_f) + \sum_{k=1}^m loss_{M^k}(W,c_k),$$

where $loss_f$ is the loss function of the fusion branch and $loss_{M^k}$ the functions associated with each modality. During the inferrence, we will use only the central branch of the fusion. The output is therefore $|\phi_f(x_i,W)-c_f\|^2$, where $\phi_f$ is the output only of the central branch.

\section{\uppercase{Experiments}}

\subsection{Dataset}

We validate our approach over several benchmark datasets portraying complex anomalous events in various scenarios involving multiple scenes captured from different angles. All datasets comprise ‘normal’ video frames for training and a combination of anomalous and
non-anomalous frames for testing. Their features are summarised in Table \ref{Tab:dataset}.

The CHUK Avenue dataset contains 16 normal videos for training and 21 videos for testing, for a total of 30,652 frames. Test videos include anomalies like the throwing of objects, walking in the wrong direction, running, and loitering. The UCSD anomaly detection dataset contains surveillance videos of pedestrian walkways. Anomalies include presence of skaters, bikers, small carts and people walking sideways in walkways. The dataset is divided into two parts: Ped1 and Ped2. Ped1 contains 34 normal video samples for training with some perspective distortion and 36 videos samples for testing. Ped2 portrays pedestrians walking parallely to the camera plane, with 16 videos for training and 12
for testing. The complex ShanghaiTech Campus dataset is specifically used to validate the robustness of the model. The corpus contains video sources from 13 different scenes, under various lighting conditions and camera angles. It has 330 video samples for training and 107 for testing, with a total of 130 abnormal events. Anomalies include complex unusual human behaviors, presence of unusual objects and movements in the wrong direction.

\begin{table*}
\centering
\begin{adjustbox}{max width=\textwidth}
\begin{tabular}{|c|c|c|c|c|c|}
\hline
\textbf{Datasets} & \textbf{Anomalous Events} & \textbf{Sources} & \textbf{Normal Frames} & \textbf{Anomalous Frames} & \textbf{Training/Testing Frames} \\
\hline
UCSD Ped1 & 40 & 1 & 9,995 & 4,005 & 6,800 / 7,200 \\
\hline
UCSD Ped2 & 12 & 1 & 2,924 & 1,636 & 2,550 / 2,010 \\
\hline
CHUK Avenue & 47 & 1 & 26,832 & 3,820 & 15,328 / 15,324 \\
\hline
ShanghaiTech & 130 & 13 & 300,308 & 17,090 & 274,515 / 42,883 \\
\hline
\end{tabular}

\end{adjustbox}
\caption{Characteristics of the datasets used in this work.}
\label{Tab:dataset}
\end{table*}

\subsection{Results \& Discussions}
We evaluate our method on two benchmark datasets. 1) The UCSD Ped2 dataset \cite{UCSD_Ped2} which contains 16 training and 12 test videos with 12 irregular events, including riding a bike and driving a vehicle. 2) The CUHK Avenue dataset \cite{CUHK_Avenue} consists of 16 training and 21 test videos with 47 abnormal events such as running and throwing stuff. 


By construction the CAE used for pretraining is symmetric. Both encoder and decoder is made of $4$ convolution layers with bias set to zero followed by Relu as the activation function. Each convolution layer is followed by a BatchNorm2d layer. The weights are initialized using Kaiming initialization. The filter are of size $3$ x $3$ with padding and stride set to $1$. We pretrain the auto-encoder for $100$ epochs and fine tune anomaly detector(encoder) for 75 epochs. We use Adam optimizer to optimize the parameters of both the CAE and the encoder. The learning rate and weight decay hyper-parameter are set to $1 \times 10^{-2}$ and $0.1$ respectively. Initially, we conducted experiments without pretraining, which led to poor outcomes. However, incorporating both pretraining and fine-tuning significantly improved the results.

The quantitative results of our approach are tabulated in Table.\ref{Tab:results}. We evaluate the algorithm using the Area under the curve metric. We use the sklearn metric ROC AUC score function to evaluate our algorithm. We find the AUC for individual videos and the average AUC for all the videos. We compare our method with other unsupervised methods such as MPPCA \cite{MPPCA}, MPPC+SFA \cite{MPPCSFA}, ConvLSTM-AE and \cite{ConvLSTM-AE}. From Table.\ref{Tab:results} it can be seen that the performance of our method is almost similar to or better than most of the methods.

\begin{figure}
	\centering
	\includegraphics[width=8cm,height=5cm]{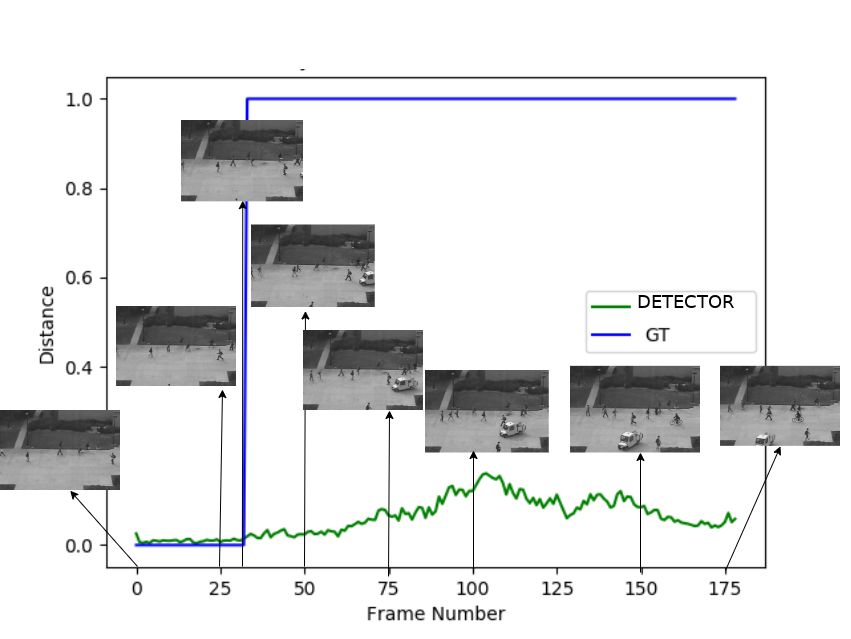}
	\caption{Anomaly detection result on Ped2 dataset, Sequence 04. GT denotes
the ground truth(Blue curve). The Green curve shows the detection from our
approach.}
 \label{fig:ped2}
\end{figure}

\begin{figure}
	\centering
	\includegraphics[width=8cm,height=5cm]{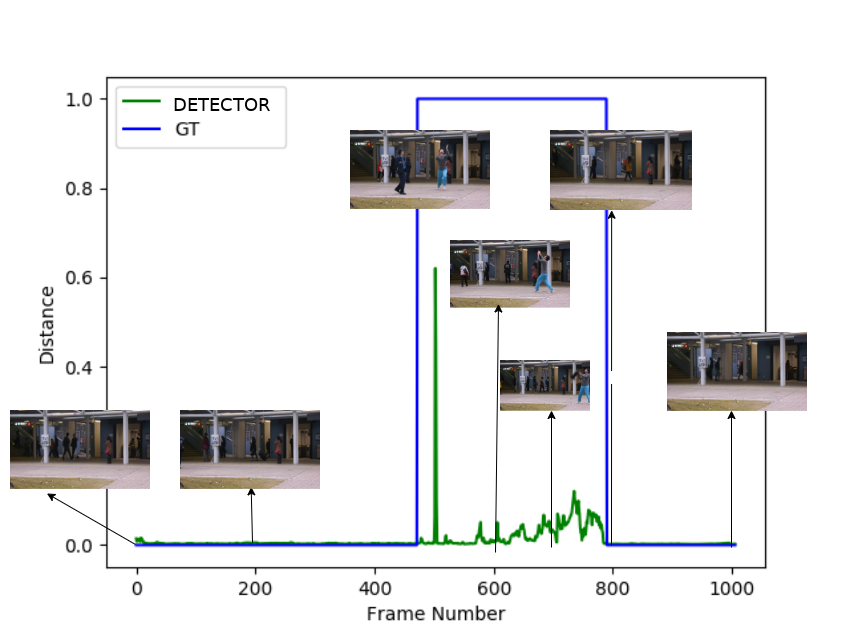}
	\caption{Anomaly detection result on Avenue dataset, Sequence 05. GT denotes
the ground truth(Blue curve). The Green curve shows the detection from our
approach}
 \label{fig:avenue}
\end{figure}

The qualitative results of our approach are shown in Figure.\ref{fig:ped2} and Figure.\ref{fig:avenue} respectively. In both the figures, the blue curve represents the ground truth data. The anomalous frame in ground truth is indicated by the value of 1 whereas the normal frames are indicated by 0 value. In Figure.\ref{fig:ped2} it can be seen that the green curve starts ascending as soon as the van enters the cameras field of view and keeps fluctuating till the end of the sequence but never drops back to the zero value. Similarly, In Figure.\ref{fig:avenue} we can see that the green curve remains zero till a person starts behaving randomly by repeatedly throwing his bag up in the air.

\begin{table}
\centering
\begin{tabular}{|c|c|c|}
\hline
\textbf{Method} & \textbf{Ped2} & \textbf{CUHK Avenue} \\
\hline
\textbf{Our Method} & 0.883 & 0.758 \\
\hline
\textbf{MPPCA } & NA & 0.69 \\
\hline
\textbf{MPPC+SFA} & 0.613 & NA \\
\hline
\textbf{ConvLSTM-AE} & 0.881 & 0.77 \\
\hline
\end{tabular}
\caption{AUC of different methodsk.}
\label{Tab:results}
\end{table}

\section{\uppercase{Conclusions \& Future Work}}
\label{sec:conclusion}

In our work we have proposed an unsupervised approach to detect anomalies in videos using a fusion of motion, depth and appearance features. Experimental evaluations on standard benchmarks demonstrate the our model performs similar to other unsupervised methods. We believe that the performance of our method can be further improved by incorporating other modalities like pose maps and audio. In our current work we just train the parameters for the fusion block. In the future we would like to simultaneously train and update the parameters of the different modalities along with the fusion block.

\bibliographystyle{apalike}
{\small
\bibliography{example}}

\end{document}